\documentclass[journal]{IEEEtran}

\usepackage{listings}

\usepackage{graphicx}
\usepackage{multirow} 
\usepackage{array}
\usepackage{overpic}

\usepackage{hyperref}
\hypersetup{hypertex=true,
            colorlinks=true,
            linkcolor=blue,
            anchorcolor=blue,
            citecolor=blue}
\usepackage{tcolorbox}

\usepackage{amsmath} 
\usepackage{amssymb}
\usepackage[T1]{fontenc}

\begin{document}

\title{Robot Detection System 2: Design of Sensor System}

\author{Jinwei~Lin,~\IEEEmembership{Member,~IEEE, ACM}
        
\thanks{Jinwei Lin is come from Monash University,} 
\thanks{ORCID of Jinwei Lin:0000 0003 0558 6699,} 
\thanks{Manuscript written in 2018.}} 

\markboth{Journal or Conference of \LaTeX\, No.~1, May~2023}
{Shell \MakeLowercase{\textit{et al.}}: Simple  Arrow Area Architecture Template}
\maketitle


\begin{abstract}
Front-following is more technically difficult to implement than the other two human following technologies, but front-following technology is more practical and can be applied in more areas to solve more practical problems. The design of sensors structure is an important part of robot detection system. In this paper, we will discuss basic and significant principles and general design idea of sensor system design of robot detction system. Besides, various of novel and special useful methods will be presented and provided. We use enough beautiful figures to display our novel design idea. Our research result is open source in 2018, and this paper is just to expand the research result propagation granularity. Abundant magic design idea are included in this paper, more idea and analyzing can sear and see other paper naming with a start of Robot Design System with Jinwei Lin, the only author of this series papers.
\end{abstract}


\begin{IEEEkeywords}
Robot, Detection, System, Novel, Sensors
\end{IEEEkeywords}


\section{Introduction}

In this project, we propose two sensor model system models, one based on a rectangular four-vertex angle model and one based on a rectangular four-sided center point model. Both modes have their own advantages and disadvantages. The model based on the rectangular four-vertice angle design has high detection precision for the detection areas at four corners, can perform double detection of a wide range, and can perform detection data superposition processing to achieve less detection error. The disadvantage of this model is that the installation structure is more complicated. The model design based on the rectangular four-sided center has the advantages of simple installation, stable scanning structure, strong front scanning capability, dual-mode data processing, and the ability to detect and track the influencing factors of the surrounding environment while the robot following target person. This scanning mode also has the advantage of performing a wide range of double detection data superposition to achieve less error. Moreover, the scanning detection accuracy of the scanning area in the four corners is also high, but not higher than the rectangular four-corner model. Due to the limitation of writing time, the project paper will first discuss and compare the differences between the two scanning modes in the later algorithm control part. Then, in the following algorithms, all of them will be explained and analyzed based on the rectangular four-sided central model. And the model is very suitable for the dual-mode scanning algorithm proposed in this paper. For other detailed analysis based on the rectangular four-corner model, please pay attention to the new or other versions of the project design paper of this paper series.

\section{Two Main Sensor Systems}

\subsection{Sensor System 1}
This section describes a sensor system based on a rectangular four-corner angle. In this mode, the robot's sensor system consists mainly of eight LRFs, abstracting the robot into a cube, and the eight LRF sensors are mounted in two corners above the cube at two or two corners. One apex angle is provided for each LRF group, and one LRF is composed of two LRFs that are in the same parallel to the Z axis. For convenience of explanation, we will give the sensor system model of the four apex angles of the robot rectangle as the four-angle sensor system model, referred to as the four-corner model. As shown in Figure\ref{fig1}:

\begin{figure}[t]
\centering
\includegraphics[width=0.99\columnwidth]{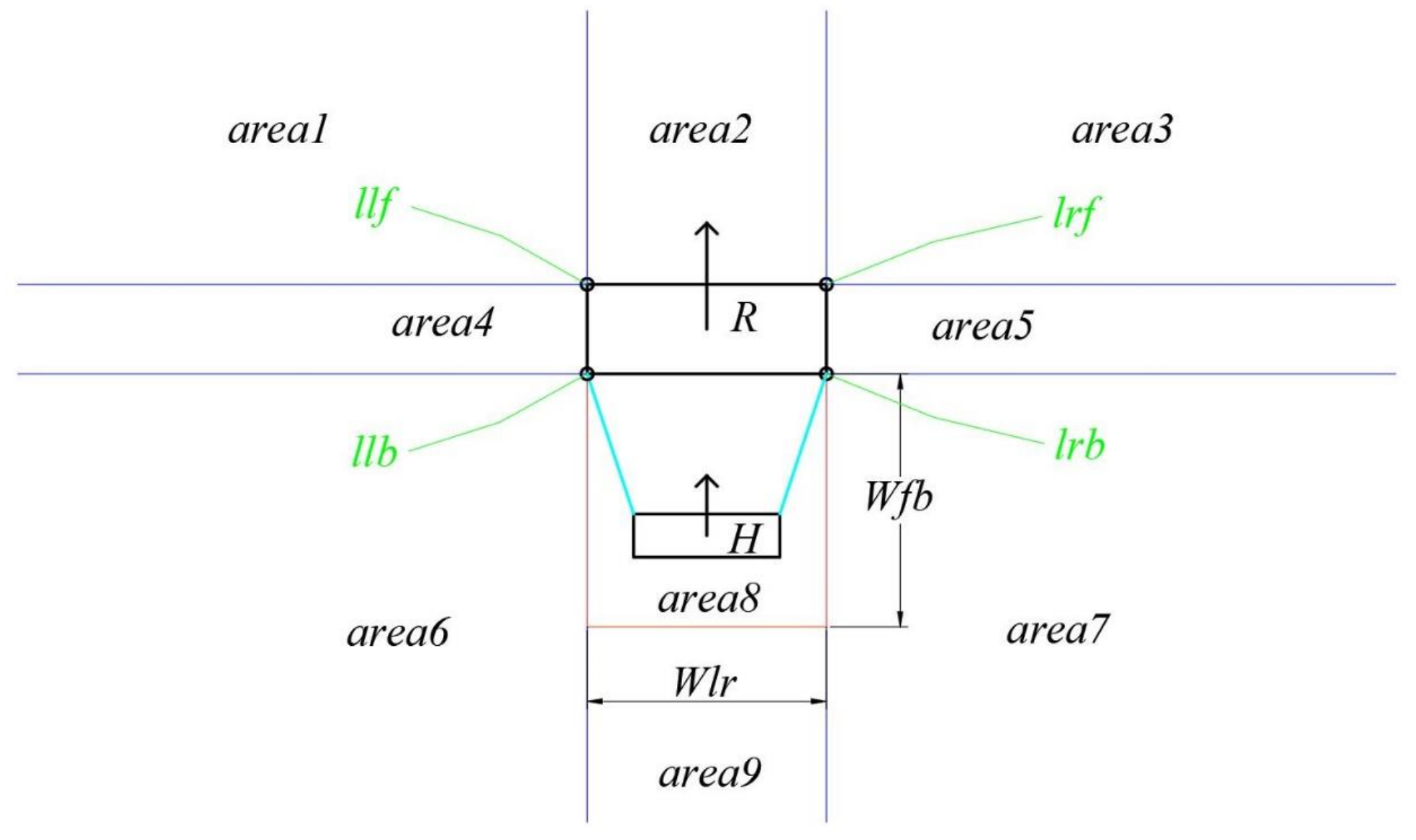}
\caption{Distribution of LRF detection areas based on four-vertex model robots.}
\label{fig1}
\end{figure}

We plot in 2D. Abstract the robot into a rectangle called a robot rectangle and mark it with R. We also abstract the target person into a rectangle called a human rectangle, marked with H. Here we define the width of the robot rectangle to be slightly larger than the human rectangle corresponding to the target person. Eight LRFs are mounted in the four corners of the robot rectangle. The two LRFs in the left front corner are labeled llf, and the two LRFs in the right front corner are labeled lrf. The two LRFs in the left rear corner are labeled llb, and the two LRFs in the right rear corner are labeled lrb. The horizontal and vertical lines have been changed in the four corners of the robot. The robot and the human space are divided into nine areas, which are marked as area1, area2, ..., area9. There is a restricted area behind the robot rectangle, which is just area8. The restricted area is a rectangle whose left to right width is equal to the width of the LRF at the back of the robot rectangle, labeled Wlr. The front and rear widths of the restricted area are marked as Wfb. The design premise of the restricted area is to be able to completely include humans in the restricted area, but to leave a certain amount of free space for human activities. Please note that the restricted area is only a virtual algorithm area, and the real human activity is not restricted by this area. The idea of a restricted area is to define a virtual area. If the target person is in this area, then the robot is locked. If the target person is not in this area, then the robot at this time is said to be in a disengaged state. Obviously, when the robot is locked, the robot shows good forward tracking to humans. When the robot is in the disengaged state, the robot cannot show good forward tracking to humans. At this time, the robot following can no longer be called front-following. In Figure\ref{fig1}, when the robot is in area8, the robot is said to be in a locked state. When the robot is in another area, the robot is said to be in a disengaged state.

The purpose of setting up the restricted area is very clear, to serve the highly accurate front-following algorithm. If a robot is always in a disengaged state in the process of following the human forward, then the robot is obviously not a good front following robot. Namely, a good robot front-following algorithm should keep the robot as locked as possible. Our robot control algorithm is based on this design.

In the above analysis, we assume that the cross-sectional area of the robot is larger than the human body’s, which is more common in the actual design of general robots. There was another situation what we can't ignore under these circumstances is that the cross-sectional area of the robot is smaller or very close to humans. In this case, the above modeling does not apply. In order to solve this problem, we propose the concept of proportional expansion area model. The purpose of the expansion area model is to make the virtual robot and the rear of the real robot on the same plane, and then construct the virtual robot, so that the cross-sectional area of the virtual robot is larger than that of the human body using the virtual robot instead of the real robot for algorithm calculation. This will facilitate the setting and operation of restricted areas, as shown in the following Figure\ref{fig2}:

\begin{figure}[t]
\centering
\includegraphics[width=0.99\columnwidth]{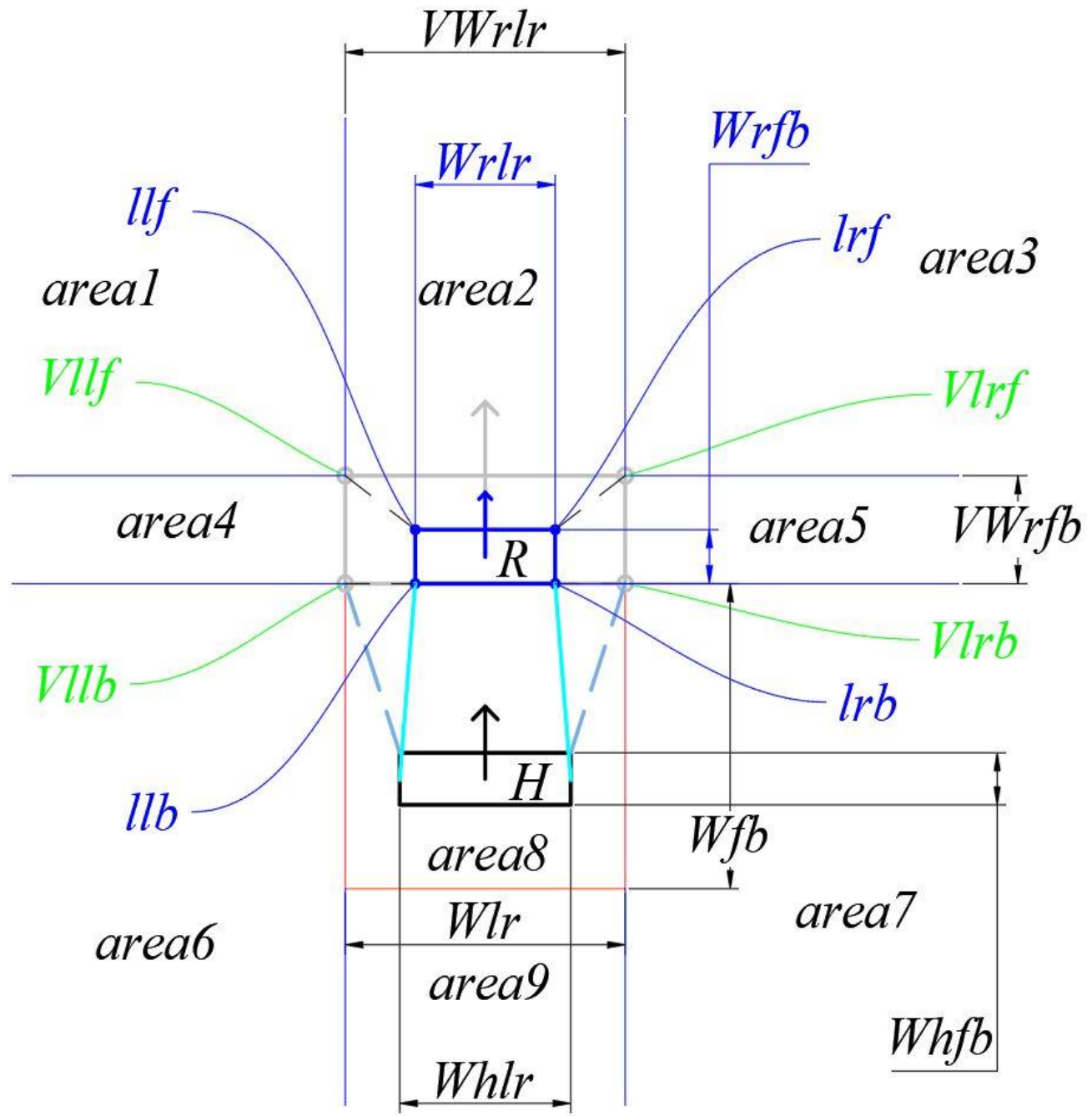}
\caption{Distribution of LRF detection areas based on four-vertex model robots.}
\label{fig2}
\end{figure}

As shown in Figure\ref{fig2}, the blue rectangle represents the real robot rectangle, marked with R. The gray rectangle represents the virtual robot rectangle. The positions of the LRFs of the two corners of the four corners of the real robot are labeled llf, llb, lrf, lrb, respectively. Let the four vertices of the virtual robot be P1, P2, P3, and P4. The four vertices defining the restricted area are P3, P4, Pa, Pb. After the scale expansion, the positions of the LRF groups in the four corners of the virtual robot rectangle are Vllf, Vllb, Vlrf, Vlrb. The front and rear width of the virtual robot rectangle is VWrfb, and the left and right width is VWrlr. It is assumed that the front and rear widths of the restricted area are Wfb, and the left and right widths are Wlr. The front and back width of the human rectangle is Whfb, and the left and right width is Whlr. The front and back width of the real robot rectangle is Wrfb, and the left and right width is Wrlr. In this case, the left and right widths of the real robot rectangle are smaller than the left and right widths of the human rectangle, that is, Wrlr < Whlr. Wfb and Wlr are automatically set up following the control mechanism. The process of setting is as follows:

First, we need to pre-set two experimental experience values, namely p and $\xi$, $\epsilon$, and specify:

\begin{equation}
\label{eq1}
\begin{cases}p=\frac{VWfb}{Wrfb}=\frac{VWlr}{Wrlr}=\frac{Wlr}{Wrlr}\\k1=\frac{Wrfb}{Whfb}\\k2=\frac{Wrlr}{Whlr}\\p>1\\p=\xi+k2\\\varepsilon=\frac{Wfb}{Wlr}\end{cases}
\end{equation}

Here $\xi$ and $\epsilon$ should be the constant ratios of the robots that are designed to be specific to a particular mechanical and circuit design. Therefore, for a structure determined robot, the values of $\xi$ and $\epsilon$ should be deterministic or not change (n.b.: the post-upgrade or modification of the robot is considered here). However, since the human $W_{hlr}$ that the robot follows each time is not necessarily the same, the corresponding p value is not necessarily the same each time the robot performs the following process.

\begin{figure}[t]
\centering
\includegraphics[width=0.99\columnwidth]{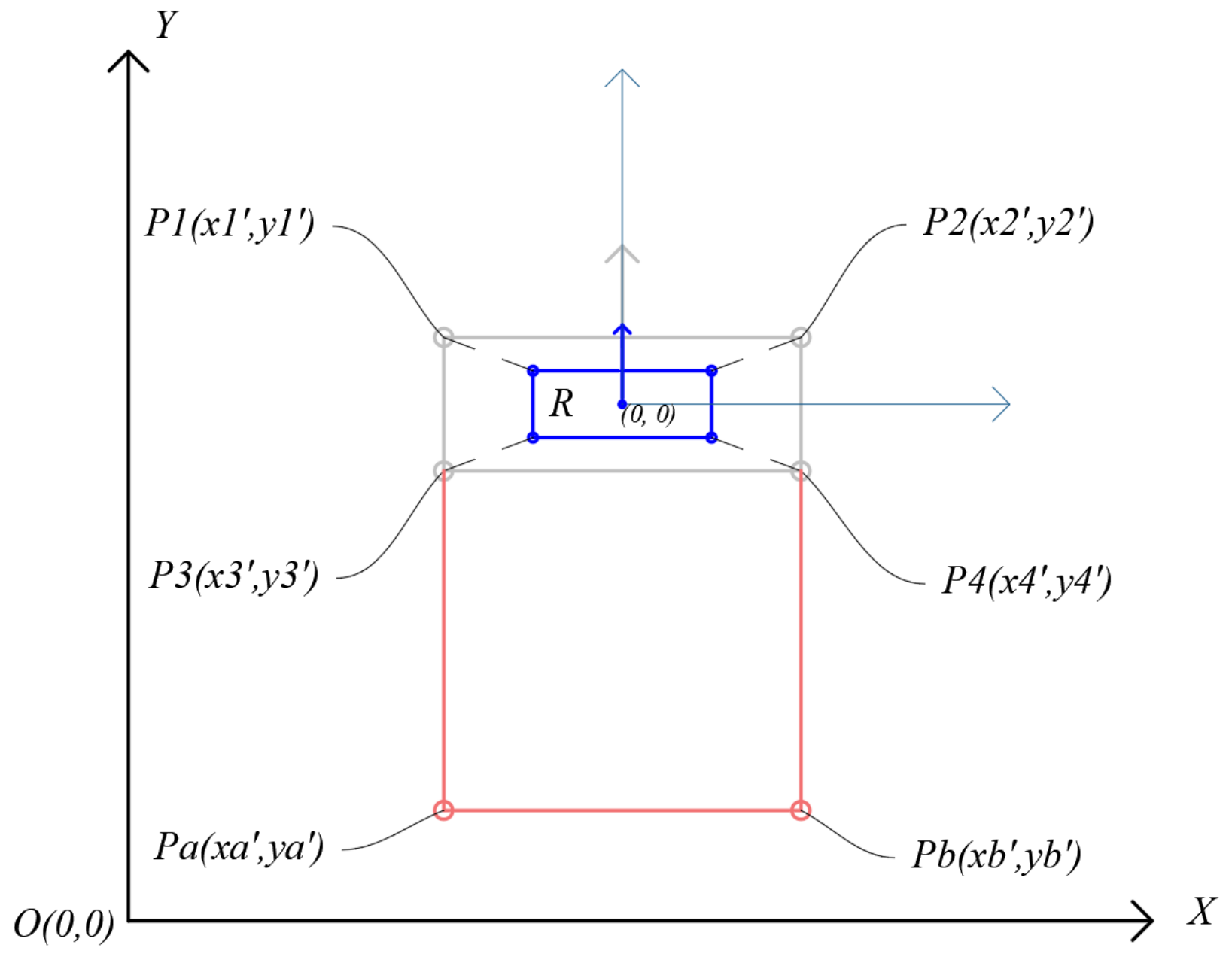}
\caption{Schematic diagram of the key points of the scale-up regional model1.}
\label{fig3}
\end{figure}

\begin{figure}[t]
\centering
\includegraphics[width=0.99\columnwidth]{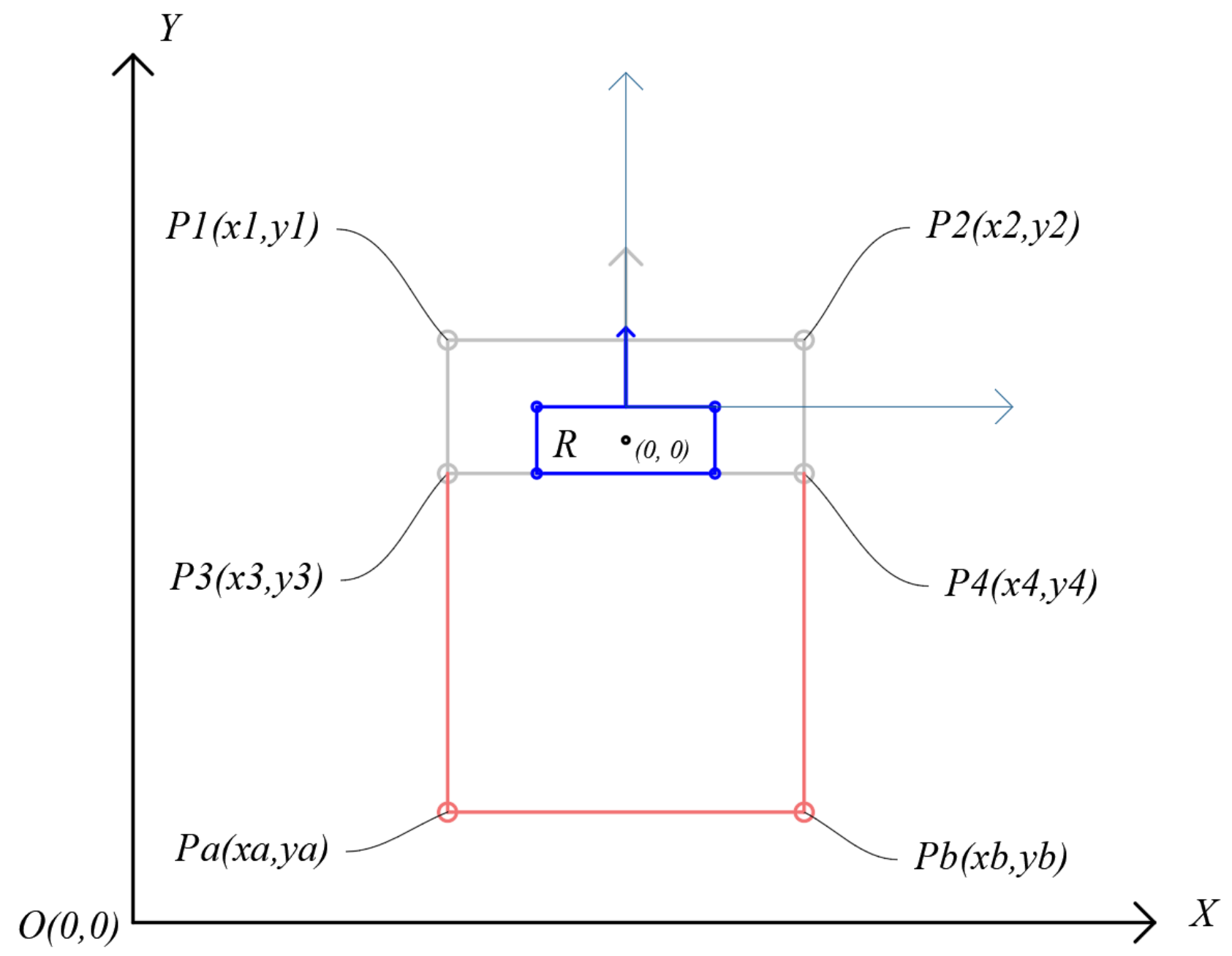}
\caption{Proportion 2 of the key points of the proportional expansion area model, moving down the origin.}
\label{fig4}
\end{figure}

In theory, p>1. The optimal range of p should be $1.2 \le p \le 2$. $K1$ is not necessarily equal to $k2$, and our focus is on $k2$. After experimental demonstration, we can determine the proportional constant of a dividing line, such as 1.2. If $k2$ satisfies $k2>1.2$, there is no need to scale up, otherwise scale expansion is required. The process of scale expansion is as follows:

1. The robot calls the sensor system to scan the left and right width  $W_{hlr}$ of the corresponding human rectangle when the human stands; 2. Find k2 from Equation\ref{eq1}, and find p in combination with the inherent $\xi$ of the robot; 3. From the Equation\ref{eq1}, combined with the inherent $\epsilon$ of the robot to find Wfb, Wlr; 4. Determine the $V_{Wfb}$ and $V_{Wlr}$ by the Equation\ref{eq1} in combination with the known $W_{rfb}$ and $W_{rlr}$ and p;5. Set the geometric center coordinates of the real robot rectangle to (0,0), as shown in Figure\ref{fig3}, then there are shown as Equation\ref{eq2}. 6. After obtaining the initial coordinates of each point, move the whole robot model in the model down by $\frac{1}{2(V_{Wrfb}-W_{rfb})}$ 1/2 units in the negative direction of the Y-axis. As shown in Figure\ref{fig4}, the final coordinates of each point are shown as Equation\ref{eq3}.

\begin{equation}
\label{eq2}
\begin{cases}
x1^{\prime}=-x2^{\prime}=x3^{\prime}=-x4^{\prime}=xa^{\prime}=-xb^{\prime}=\\-\frac{1}{2}VWlr=-\frac{1}{2}Wlr\\y1^{\prime}=y2^{\prime}=\frac{1}{2}VWfb=\frac{p}{2}Wfb\\y3^{\prime}=y4^{\prime}=-\frac{1}{2}VWfb=-\frac{p}{2}Wfb\\ya^{\prime}=yb^{\prime}=-(Wfb+\frac{1}{2}VWrfb)=\\-(\varepsilon\cdot Wlr+\frac{1}{2}VWrfb)
\end{cases}
\end{equation}

Please note that in general, the distance between the two LRFs in the LRF group in the two groups is very close on the Z axis. In general, we can ignore the height distance difference between them. However, in the pursuit of high-precision algorithms, it is also necessary to consider the difference in height distance between two groups of LRFs. The specific part will be covered in the algorithm analysis section.

\begin{equation}
\label{eq3}
\begin{cases}
x1=x3=xa=x1'=x3'=xa'\\x2=x4=xb=x2'=x4'=xb'\\y1=y2=y1'+\frac12(VWrfb-Wrfb)\\y3=y4=y3'+\frac12(VWrfb-Wrfb)\\ya=yb=ya'+\frac12(VWrfb-Wrfb)
\end{cases}
\end{equation}

Now, we can determine the coordinates of each vertex of the restricted area and the virtual robot rectangle. And we can determine the extent and size of the virtual robot rectangle and its corresponding restricted area. From the above analysis, we know that only the restricted area corresponding to the real robot rectangle can not satisfy the case of including the human rectangle while retaining a certain space, and then it is necessary to scale the real robot rectangle.

After determining the size of virtual robots and restricted areas, an obvious problem will be: if our robot only have four LRF groups distributed in four corners, how can we assign working areas to these four LRF groups to enable them to effectively develop their due detection capabilities? To solve this problem, we propose a 45-degree partition method. The specific implementation process of the 45-degree division method is as follows:

1. Using the four vertices of the virtual robot rectangle as the starting point, make a ray with an angle of 45 degrees from the left and right horizontal boundary lines;

\begin{figure}[t]
\centering
\includegraphics[width=0.99\columnwidth]{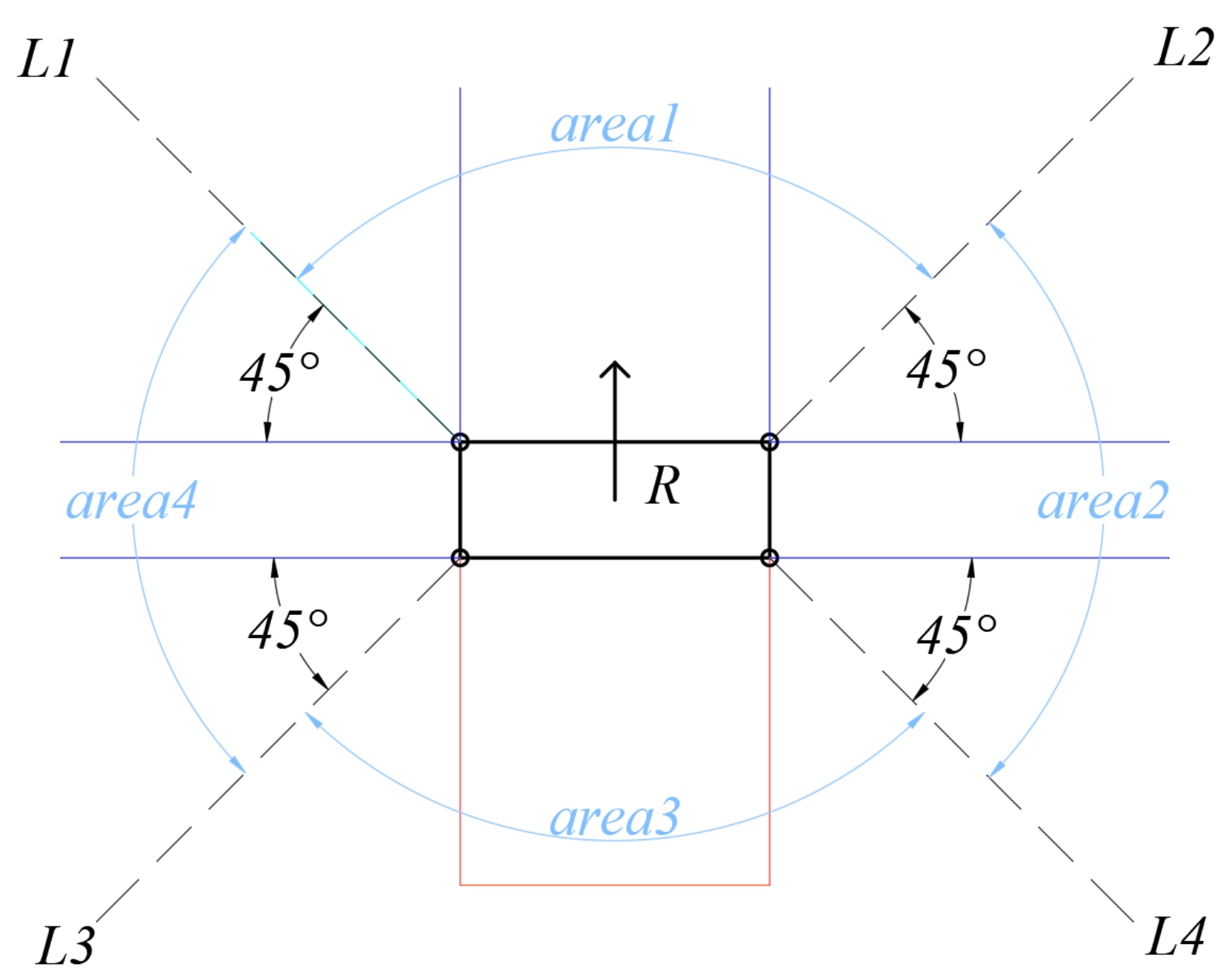}
\caption{45-degree partitioning method based on two-dimensional space.}
\label{fig5}
\end{figure}

2. As shown in Figure\ref{fig5}, the area where the robot rectangle is located is divided into four areas according to the two-dimensional space: area1, area2, area3, area4. Then, two LRFs in each LRF group are allocated into two adjacent regions. For example, one LRF partition in the LRF group at point P1 in FIGS. 3 and 4 is specified to detect area1, and another LRF partition is used to detect area4. A similar partition designation is made for the LRF groups at the remaining three vertices. The final result of the division is that each detection area is checked by two LRFs.

3. As shown in Figure 6, assume that there are 8 points to be measured in the space, which are represented by P1, P2, ..., P8. Each checkpoint is subject to the following rules and definitions:
3.1. If the point to be measured is in area1, the two LRFs in front of the virtual robot are responsible for the detection. In fact, it is detected by the two LRFs in front of the real robot.
3.2. If the point to be measured is in area2, the two LRFs on the right side of the virtual robot are responsible for the detection. In fact, the two LRFs on the right side of the real robot are responsible for the detection.
3.3. If the point to be measured is in area3, the two LRFs behind the virtual robot are responsible for the detection. In fact, it is detected by the two LRFs behind the real robot.
3.4. If the point to be measured is in area4, the two LRFs on the left side of the virtual robot are responsible for the detection. In fact, it is detected by the two LRFs on the left side of the real robot.
3.5. With this division method, a special type of point will appear: a point (eg, P1) on the 45-degree dividing line, assuming that the detected object represented by the point (eg, P1) is just entering the detection scanning area. And at the moment of scanning, it is just on a certain 45 degree dividing line. When the robot detection system allocates LRFs for such points, it will drive the two LRFs of the same LRF group corresponding to the point to be jointly responsible for detection.
3.6. For those points located on the horizontal dividing line or vertical dividing line of the virtual robot rectangle, such as P8, we will ignore the influence of the horizontal dividing line or vertical dividing line of the virtual robot rectangle, and directly distribute the points according to the corresponding detection area. (area4) The two LRFs assigned to the left of the robot are responsible for the detection. For those points that are very close to the 45-degree dividing line, the LRF groups are still strictly distributed to detect those points according to the area where the center of the points are located and the division principle. For example, point P6 will be divided into area2, that is, the two LRFs on the right side of the robot are responsible for the detection.

\begin{figure}[t]
\centering
\includegraphics[width=0.99\columnwidth]{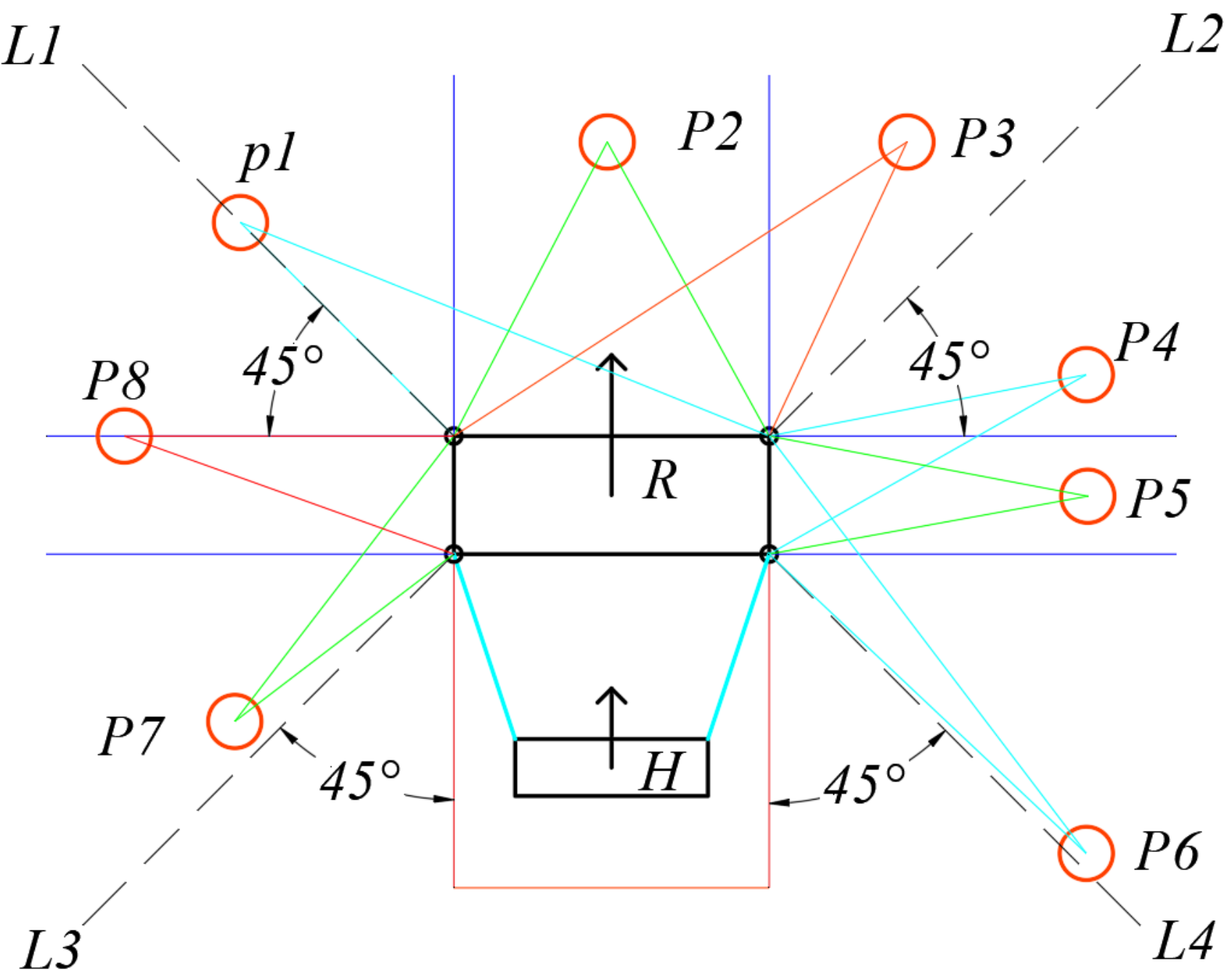}
\caption{Distribution of LRF detection regions using 45-degree region partition method.}
\label{fig6}
\end{figure}

\subsection{Sensor System 2}
In this part, we will introduce a new sensor system. The system is also composed of eight LRFs, and each two LRFs constitute one LRF group, and then four LRF groups are respectively installed at the center of the four sides of the robot rectangle. Although in subjective feeling, the robot only changed in the position of the four LRF groups compared to the previous sensor system, the two sensor systems will make a huge difference in algorithm characteristics and performance characteristics. For the sake of convenience, we refer to this sensor model based on the uniform distribution of the four-sided center as the four-sided central model.

\begin{figure}[t]
\centering
\includegraphics[width=0.99\columnwidth]{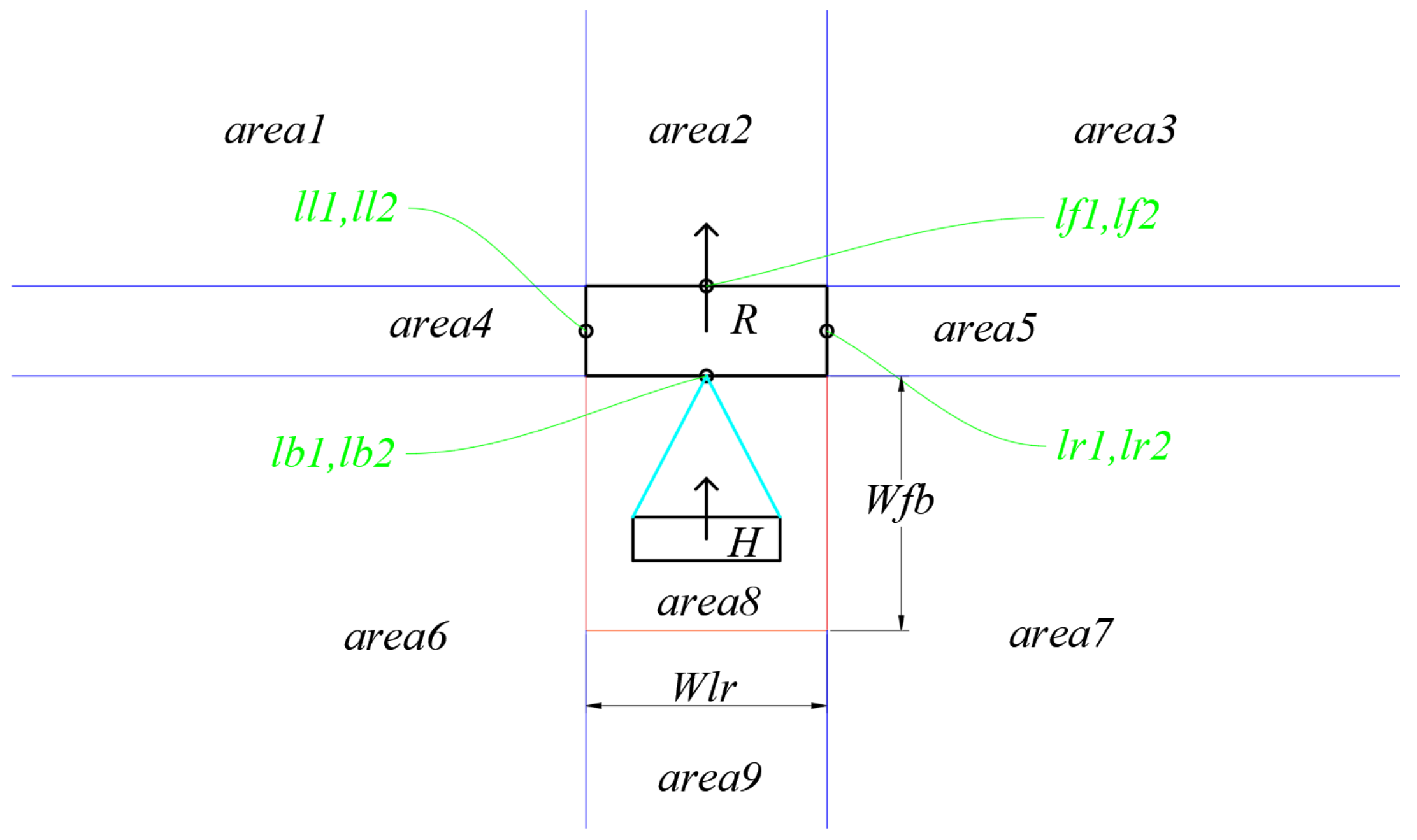}
\caption{Robotic sensor system based on rectangular four-sided center model.}
\label{fig7}
\end{figure}

As shown in Figure\ref{fig7}, similarly, we make an infinitely extended straight line along the four sides of the robot rectangle, and divide the two-dimensional space corresponding to the space in which the robot is located into eight regions. Then, the restricted area is introduced again, and the corresponding two-dimensional space is divided into nine areas in total.

Then, four LRF groups are respectively disposed at the center of the four sides of the robot rectangle, and each LRF group is composed of two LRFs in the same linear direction parallel to the Z axis. The four LRF groups are respectively recorded as: (ll1, ll2), (lf1, lf2), (lr1, lr2), (lb1, lb2).

\begin{figure}[t]
\centering
\includegraphics[width=0.99\columnwidth]{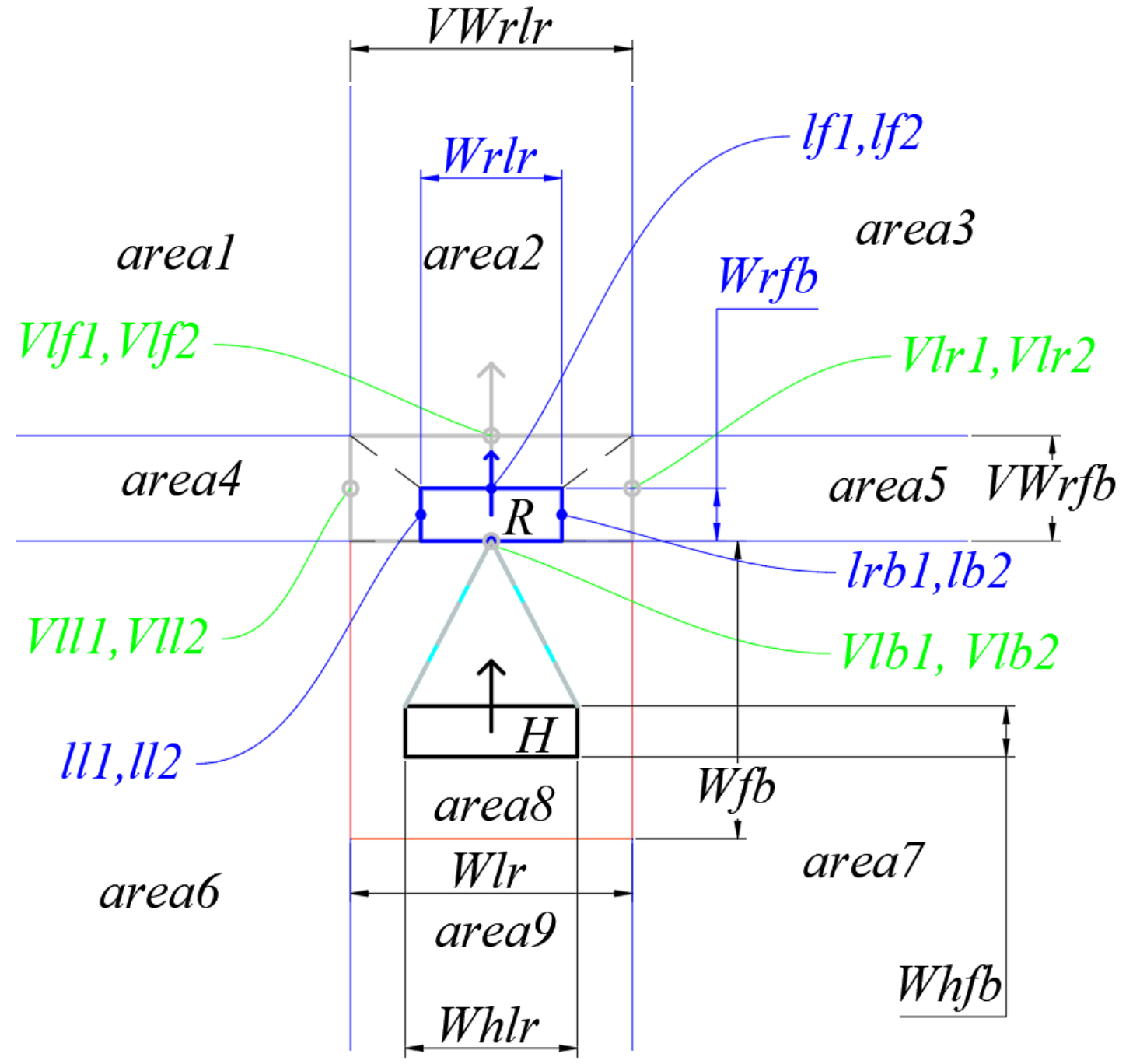}
\caption{Robotic sensor system based on rectangular four-sided center model.}
\label{fig8}
\end{figure}

As shown in Figure\ref{fig8}, in terms of parameter configuration, in addition to the positional parameters of the sensor, other model parameters based on the rectangular four-sided center model are the same as the four-vertex model described above. Therefore, when defining the algorithm parameters of the sensor model in the center of the four sides, we only need to redefine the position parameters of the sensor, and the other model parameters will remain unchanged. We set the LRF group (Vlf1, Vlf2) at the center point of the front edge of the virtual robot rectangle, set the LRF group (Vlb1, Vlb2) at the center point of the rear edge, and set the LRF group (Vll1) at the center point of the left edge. Vll2), set the LRF group (Vlr1, Vlr2) at the center point of its right edge. We set the LRF group (lf1, lf2) at the center point of the front edge of the real robot rectangle, set the LRF group (lb1, lb2) at the center point of the rear edge, and set the LRF group (ll1) at the center point of the left edge. Ll2), set the LRF group (lr1, lr2) at the center point of its right edge.

After defining the parameters, we also introduce a scale expansion mechanism for the four-sided central model, so that the four-sided central model’s design methods and ideas can be applied to robots of different body types and structural designs.

Similarly, after scale expansion, we define the LRF groups located in the four corners of the virtual robot at the positions Vllf, Vllb, Vlrf, Vlrb. The front and rear width of the virtual robot is VWrfb, and the left and right width is VWrlr. It is assumed that the front and rear width of the restricted area is Wfb, and the left and right width is Wlr. The front and back width of the human rectangle is Whfb, and the left and right width is Whlr. The front and back width of the real robot rectangle is Wrfb, and the left and right width is Wrlr. In the case of proportional expansion, the left and right width of the real robot rectangle should be less than the left and right width of the human rectangle, that is, Wrlr<Whlr. The values of Wfb and Wlr are automatically set by the follow control mechanism. The corresponding setting process is as follows:

First, we need to pre-set three experimental experience values, defined as p and $xi$, $\epsilon$, and specify: p and $xi$, $\epsilon$ satisfy Equation\ref{eq1}. Similarly, $xi$ and $\epsilon$ should be the constant ratios obtained after many experiments, which are suitable for the robot group of a same particular mechanical and circuit structure. Therefore, for a group of robots based on a same particular structure, the values of $xi$ and $\epsilon$ should be deterministic. However, since the Whlr of the human being followed by the robot is not necessarily the same each time in following the task, the p values corresponding to the robots in the different following tasks are not necessarily the same.

\begin{figure}[t]
\centering
\includegraphics[width=0.99\columnwidth]{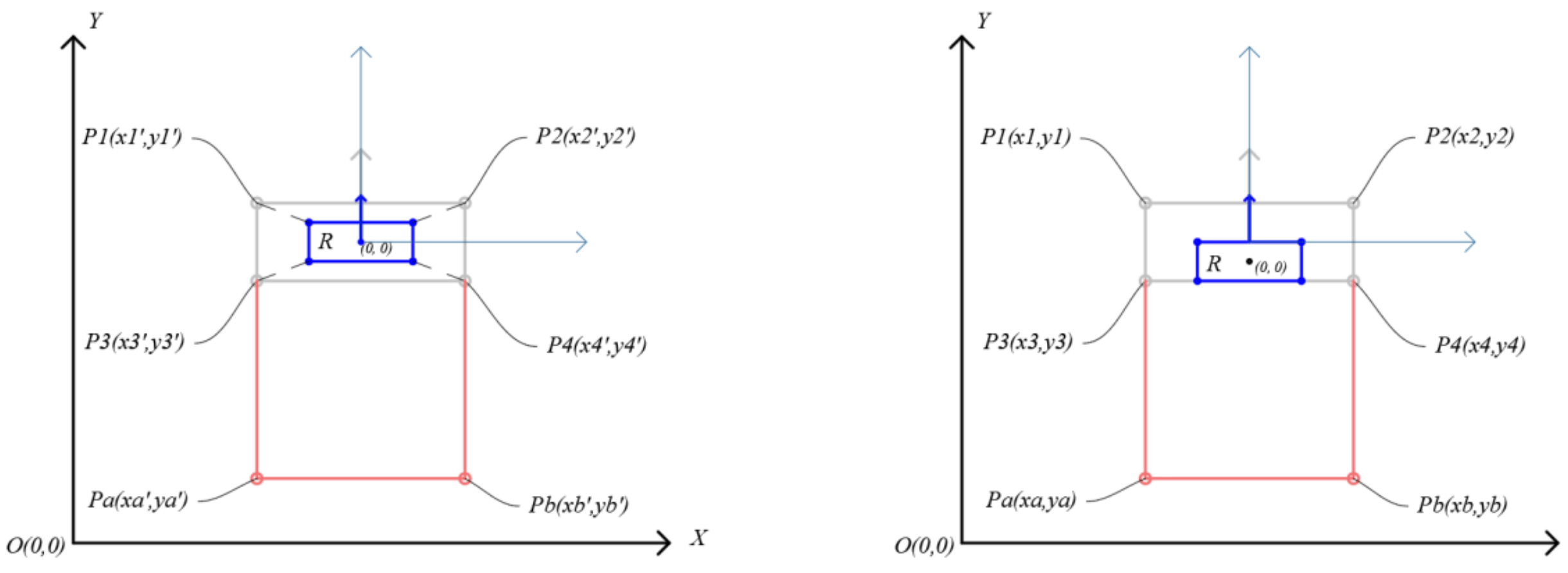}
\caption{Proportional expansion mechanism of the four-sided central model.}
\label{fig9}
\end{figure}

In theory, $p>1$. The optimal range of P should be $1.2\le p \le 2$. K1 is not necessarily equal to k2, and our focus is on k2. If k2 satisfies $k2 > 1.2$, there is no need to do the scale expansion, otherwise scale expansion is required. The process of scale expansion is as follows: 1. First, the robot calls the sensor system to scan the corresponding left and right widths Whlr when the human is standing; 2.Find k2 from Equation\ref{eq1}, and find p in combination with the $xi$ parameter value inherent in the robot; 3. From the Equation\ref{eq1}, combined with the inherent $\epsilon$ parameter value of the robot to find Wfb, Wlr; 4. Determine the VWfb and VWlr by the Equation\ref{eq1} in combination with the known Wrfb and Wrlr and p; 5. Set the coordinates of the geometric center of the real robot rectangle to (0,0), 
as shown in the left figure of Figure\ref{fig9}, there are shown in Equation\ref{eq4}. 6. After obtaining the initial coordinates of each point, move the rectangular model of the robot in the model downward by $\frac12(VWrfb-Wrfb)$ units in the negative direction of the Y-axis, as shown in the right figure of Figure 9. The final coordinates of the point are shown as Equation\ref{eq5}.

\begin{equation}
\label{eq4}
\begin{cases}
x1'=x3'=0\\y1'=-y3'=\frac12VWrfb\\x2'=-x4'=xb'=-xa'=\frac12VWrlr\\y2'=y4'=0\\ya'=yb'=-(\frac12VWfb+Wfb)
\end{cases}
\end{equation}

\begin{equation}
\label{eq5}
\begin{cases}
x1=x3=0\\x2=-x4=xb'=-xa'=\frac{1}{2}VWrlr\\y1=\frac{1}{2}VWrfb+\frac{1}{2}(VWrfb-Wrfb)\\y3=-\frac{1}{2}VWrfb+\frac{1}{2}(VWrfb-Wrfb)\\y2=y4=0+\frac{1}{2}(VWrfb-Wrfb)\\ya=yb=-\left(\frac{1}{2}VWfb+Wfb\right)+\frac{1}{2}(VWrfb-Wrfb)
\end{cases}
\end{equation}

With simplification, the above expressions of Equation\ref{eq5} can be shown as Equation\ref{eq6}:

\begin{equation}
\label{eq6}
\begin{cases}
x1=x3=0\\x2=-x4=xb^{\prime}=-xa^{\prime}=\frac12VWrlr\\y1=VWrfb-\frac12Wrfb\\y3=-Wrfb\\y2=y4=\frac12(VWrfb-Wrfb)\\ya=yb=-\left(\frac12Wrfb+Wfb\right)
\end{cases}
\end{equation}

\begin{figure}[t]
\centering
\includegraphics[width=0.99\columnwidth]{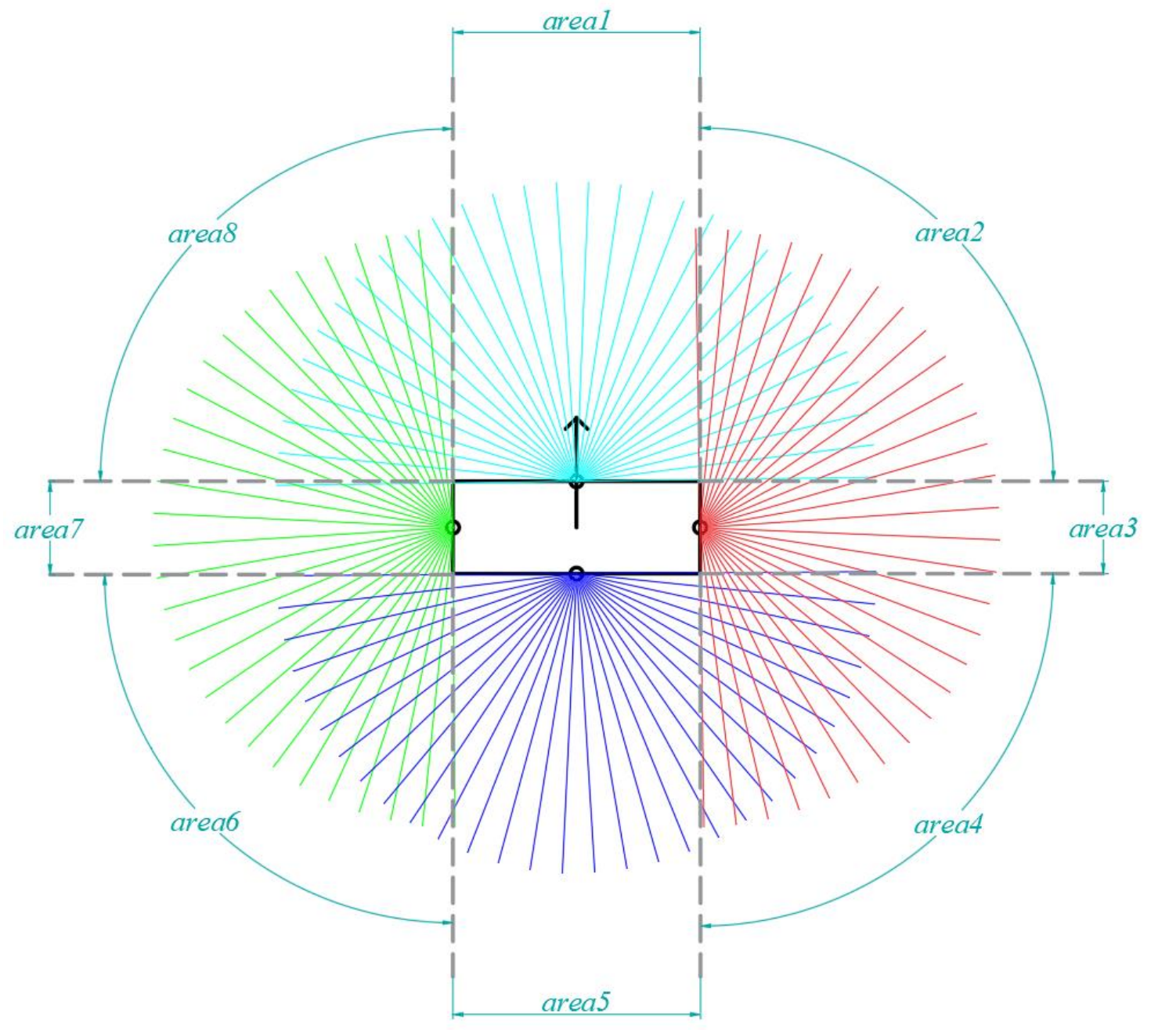}
\caption{Configuration of the detection area of the four-sided center model.}
\label{fig10}
\end{figure}

Next, we will assign the corresponding detection area to the four-sided central model. As shown in Figure 10, we make an infinitely extended parallel line along the four sides of the robot rectangle, and divide the two-dimensional space corresponding to the space in which the robot rectangle is located into eight detection area sections. The eight detection area sections are labeled as area1, area2, ..., area8, respectively. The eight detection areas can be further divided into two categories, one is an overlapping detection area, and the other is a separate detection area. The overlap detection area refers to: the detection area can be detected by the central LRF group of the adjacent two sides of the robot rectangle, such as area2, area4, area6, area8 in Figure\ref{fig10}. The separate detection zone refers to an area that can only be detected by the LRF group at the center of one side of the robot rectangle, such as area1, area3, area5, area7 in Figure\ref{fig10}. For example, the overlap detection area area8 will be detected by the left and front LRF groups of the robot rectangle, and the area2 will be detected by the front and right LRF groups of the robot rectangle. The area4 is detected by the right and rear LRF groups of the robot rectangle, and the area6 is detected by the left and rear LRF groups of the robot rectangle. Area1 is detected by the LRF group in front of the robot rectangle, area3 is detected by the LRF group on the right side of the robot rectangle, area5 is detected by the LRF group behind the robot rectangle, and area7 is detected by the LRF group on the left side of the robot rectangle. (N.B.: There is a black forest illusion in Figure\ref{fig10}, such that the extensions of the corresponding parallel sides of the robot rectangle do not appear to be parallel, but in reality the two parallel lines are parallel).

The advantage of using a four-sided central model for sensor system configuration is that we can completely and seamlessly cover the space in which the robot is located, without the occurrence of a missing detection area. Moreover, in Front-Following, in reality, the four regions corresponding to the front, the back, the left, and the right of the robot are less likely to be disturbed by the outside than the four diagonal regions of the rectangle. In the following follow-up system algorithm design, we will mention the importance of the detection capability of the LRF groups in the diagonal detection areas(i.e., overlapping detection areas) when the robot performs curve motion following. The overlap detection areas are detected by the dual LRF groups with respect to the individual detection areas, so the detection accuracy of the overlap detection areas is higher than that of the individual detection areas. Through the four-sided center model, we can also find that the areas of the overlapping detection areas are much larger than the areas of the individual detection areas.

\begin{figure}[t]
\centering
\includegraphics[width=0.99\columnwidth]{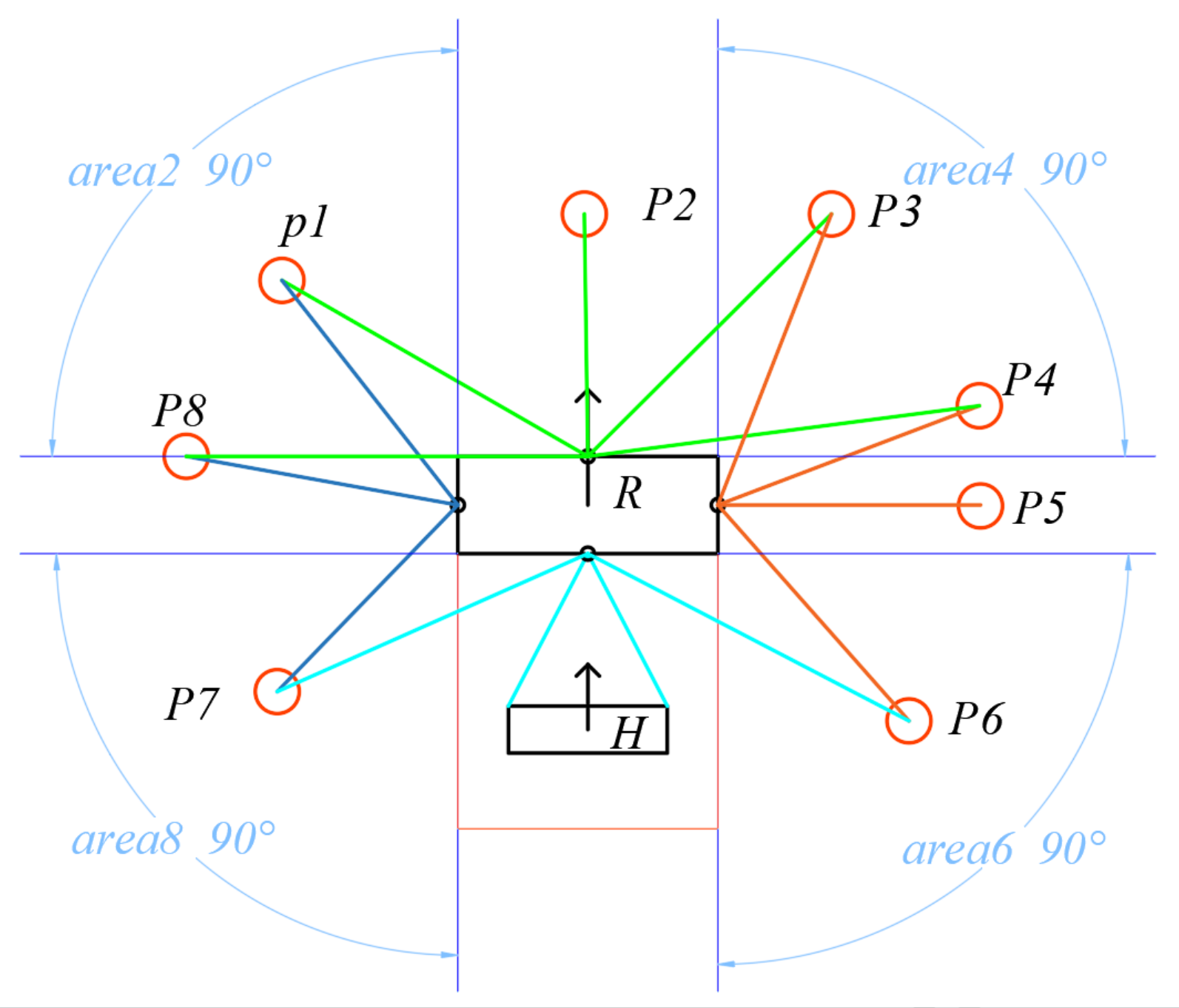}
\caption{Schematic diagram of double detection of the four-sided center model.}
\label{fig11}
\end{figure}

For example, as shown in Figure\ref{fig11}, P3, P4, P6, and P7 of the overlap detection area are both double-detected, and their detection accuracy will be higher than other points.

Figure\ref{fig11} also shows a special kind of point distribution, which is the point on the dividing line, such as P8. Whether the point on the dividing line is double-detected depends on the mechanical structure of the robot's sensor. In general, the demarcation point is also double-detected. So we also count P8 into the category of double detection points.

\section{Conclusion of Sensor System}
For content arrangement of this paper series, we make the first simple conclusion in this secession. In this paper, we made detailed analysis and discussion of two senor system implementations. For more detailed technologies analyzes, note the other papers of this paper series. Sensor system design is significant for robot detection system design, for example, combine with the area drawing detection technology \cite{lin2022multi}, can improve the detection function of robot detection system.

\bibliographystyle{IEEEtran}
\bibliography{ref}{}
\end{document}